\pdfoutput=1

\documentclass[11pt]{article}

\usepackage[final]{acl}

\usepackage{times}
\usepackage{latexsym}
\usepackage[ruled,linesnumbered]{algorithm2e}
\usepackage{algpseudocode} 
\usepackage{xcolor}

\usepackage[T1]{fontenc}

\usepackage[utf8]{inputenc}

\usepackage{microtype}

\usepackage{inconsolata}

\usepackage{colortbl} 
\usepackage{xcolor} 
\usepackage{graphicx}
\usepackage{multirow}
\usepackage{multicol}
\usepackage{booktabs}
\usepackage{amssymb}
\usepackage{amsmath}
\usepackage{xspace}
\usepackage{enumitem}
\usepackage{amssymb}
\usepackage{makecell}
\usepackage{xcolor} 
\usepackage{url}

\newcommand{\paratitle}[1]{\vspace{1.5ex}\noindent\textbf{#1}}
\newcommand{\ie}{\emph{i.e.,}\xspace}

\newcommand{\eg}{\emph{e.g.,}\xspace}

\newcommand{\ignore}[1]{}

\usepackage{tcolorbox}
\definecolor{darkorange}{RGB}{255, 140, 0}
\definecolor{lightgreen}{RGB}{145, 204, 117}
\definecolor{lightyellow}{RGB}{250, 200, 88}
\definecolor{lightred}{RGB}{238, 102, 102}
\definecolor{lightblue}{RGB}{115, 192, 222}
\newtcolorbox{promptbox}[2][Prompt]{
colback=black!5!white,
arc=5pt, 
boxrule=0.5pt,
fonttitle=\bfseries,
title=#1, 
before upper={\footnotesize}, fontupper=\fontfamily{ptm}\selectfont,
colframe=#2, 
}

\title{Sticker-TTS: Learn to Utilize Historical Experience with a \\Sticker-driven Test-Time Scaling Framework}

\author{
Jie Chen\textsuperscript{\rm{1}}$^{*}$,
Jinhao Jiang\textsuperscript{\rm{1}}$^{*}$, 
Yingqian Min\textsuperscript{\rm{1}}\thanks{Equal contribution},
Zican Dong\textsuperscript{\rm{1}},
Shijie Wang\textsuperscript{\rm{2}}, \\
\textbf{Wayne Xin Zhao}\textsuperscript{\rm{1}}\thanks{Corresponding author}
and \textbf{Ji-Rong Wen}\textsuperscript{\rm{1}} \\
\textsuperscript{1}Gaoling School of Artificial Intelligence, Renmin University of China \\
\textsuperscript{2}Northeastern University at Qinhuangdao \\
\texttt{\{ptyzchenjie,batmanfly\}@gmail.com}
}

\begin{document}
\maketitle
\begin{abstract}
Large reasoning models (LRMs) have exhibited strong performance on complex reasoning tasks, with further gains achievable through increased computational budgets at inference. However, current test-time scaling methods predominantly rely on redundant sampling, ignoring the historical experience utilization, thereby limiting computational efficiency. To overcome this limitation, we propose \textbf{Sticker-TTS}, a novel test-time scaling framework that coordinates three collaborative LRMs to iteratively explore and refine solutions guided by historical attempts. At the core of our framework are distilled key conditions—termed \textit{stickers}—which drive the extraction, refinement, and reuse of critical information across multiple rounds of reasoning. To further enhance the efficiency and performance of our framework, we introduce a two-stage optimization strategy that combines imitation learning with self-improvement, enabling progressive refinement. Extensive evaluations on three challenging mathematical reasoning benchmarks, including AIME-24, AIME-25, and OlymMATH, demonstrate that Sticker-TTS consistently surpasses strong baselines, including self-consistency and advanced reinforcement learning approaches, under comparable inference budgets. These results highlight the effectiveness of sticker-guided historical experience utilization.
Our code and data are available at \url{https://github.com/RUCAIBox/Sticker-TTS}.
\end{abstract}

\section{Introduction}

Recent advancements in foundation models, particularly when combined with reinforcement learning (RL) during training, have significantly improved the capabilities of LRMs on complex inference tasks~\cite{team2025kimi, guo2025deepseek, yang2025qwen3, zhao2025surveylargelanguagemodels}. Empirical studies demonstrate that increasing computational budgets during both training and inference phases yields consistent gains in reasoning performance. For example, OpenAI's reasoning series models (\eg o1 and o3) highlight how test-time scaling can further boost accuracy on challenging benchmarks~\cite{openai2024a,openai2024b,openai2025o3}. Unlike RL-based optimization—which incurs substantial computational overhead—test-time scaling offers a more affordable alternative, attracting growing interest for its favorable cost–performance trade-off~\cite{chen2024alphamath, Kang2024mindstar,aot-fengwei}.

Existing researches mainly propose two lines of approaches for achieving test-time scaling. A common approach executes multiple independent single-round inferences and selects the final answer via majority vote~\cite{wang2022self}. Despite its simplicity and robustness as a strong baseline~\cite{jiang2024enhancingllmreasoningrewardguided}, this strategy treats each inference as isolated, often resulting in redundant or uninformative computations. To address this limitation, recent studies have proposed an iterative multi-round inference method, where the model incorporates prior reasoning traces or final answers into subsequent inference inputs~\cite{chen2025empirical}. While this paradigm encourages history-aware reasoning, it introduces new challenges: overly verbose reasoning histories in the input may lead models to forget or overlook salient facts, and the brevity of final answers makes it difficult for models to revise earlier outputs, even when faced with inconsistencies or superior alternatives. These issues become increasingly pronounced as reasoning chains grow in length and complexity.

\begin{figure*}[t]  
\centering
\small
\includegraphics[width=0.95\textwidth]{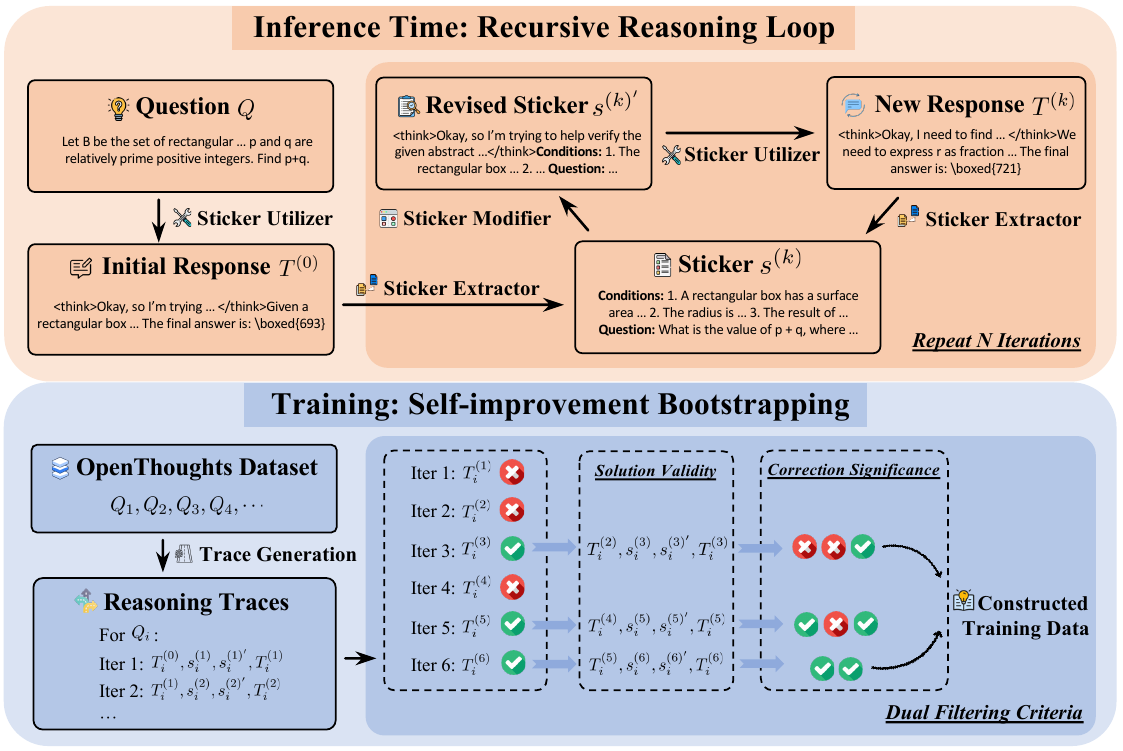}
\caption{The overall framework of our proposed Sticker-TTS.}
\label{fig:pipeline}
\vspace{-0.2cm}
\end{figure*}

To address the aforementioned challenges, we propose a novel framework aimed at striking a balance between overly verbose reasoning traces and excessively concise final answers, thereby encouraging LRMs to explore novel solution paths by leveraging historical attempts. Inspired by how humans approach long-form generative tasks—such as writing—by distilling key intermediate ideas, conclusions, or inflection points to scaffold the final output, we introduce a method to distill a compact set of essential solution cues from the lengthy reasoning processes, which we term ``\textit{stickers}''. Stickers encapsulate key conceptual anchors that guide future reasoning. At each round, we extract and refine a sticker from the previous long-form reasoning trace and embed it into the subsequent input. This approach encourages LRMs to explore alternative solutions while effectively leveraging past attempts. By using stickers as lightweight, expressive intermediates, our method enhances both reasoning robustness and efficiency.

In this paper, we propose \textbf{Sticker-TTS}, a collaborative framework designed for test-time scaling with multiple LRMs, enabling effective utilization of historical experience. The framework comprises three key components: a \textit{Sticker Extractor}, which distills concise and relevant insights (\textit{``stickers''}) from previous reasoning traces; a \textit{Sticker Modifier}, which adapts these stickers to the current context; and a \textit{Sticker Utilizer}, which integrates them to guide the model towards more effective solution strategies. During inference, these components operate iteratively, allowing the model to synthesize prior knowledge with new reasoning paths. To enhance the utility of this collaborative process, we propose a two-stage training paradigm combining knowledge distillation and self-improvement. Initially, the extractor and modifier are trained on approximately 1K distilled examples. The full framework is then used to sample collaborative reasoning trajectories, which in turn serve as new training data to iteratively refine the modules. This cycle of generation and retraining progressively enhances the model’s reasoning ability, demonstrating the promise of sticker-based collaboration for scaling test-time inference.

To validate the effectiveness of Sticker-TTS, we evaluate our method on several challenging reasoning math benchmarks, including the 2024 and 2025 AIME problem sets and OlymMATH, a recently introduced Olympiad-level math benchmark. Our framework consistently outperforms competitive baselines on both benchmarks under comparable compute bugets. For example, our method achieves a 12.42\% relative improvement over self-consistency on the AIME-25 using a 7B model. On the other hand, compared to models trained with reinforcement learning, our approach performs comparably or even better across multiple benchmarks—for instance, achieving a 9.79\% relative improvement over Skywork-OR1 on OlymMath. Moreover, when scaling computation through multi-round reasoning, our method demonstrates further performance gains, delivering an 18.75\% relative improvement over Light-R1 on AIME-25. This demonstrates the efficiency of our approach in utilizing historical experience for better test-time scaling.






\section{Method}
\label{sec:method}
Unlike traditional test-time scaling methods, our approach focuses on refining and utilizing historical experiences. We provide an overview of our method in Section~\ref{sec:overview}. {Furthermore, we introduce the inference and training of our approaches in Section~\ref{sec:recursive_reasoning_loop} and Section~\ref{sec:training}, respectively.

\subsection{Overview}
\label{sec:overview}
Our Sticker-TTS framework comprises three interrelated models: the Sticker Extractor $E$, the Sticker Modifier $M$, and the Sticker Utilizer $U$. These models work collaboratively through an iterative reasoning process. Given a reasoning trace, the Sticker Extractor first extracts and summarizes key reasoning steps and global strategies into a structured \textit{\textbf{``sticker''}}. The Sticker Modifier subsequently inspects this sticker for any mistakes, applying necessary corrections. Finally, the Sticker Utilizer generates an enhanced reasoning trace by integrating the modified sticker with the original question and the previous trace. We show the overall procedure in Figure~\ref{fig:pipeline}.

\begin{algorithm}[h]
\small
\SetKwInOut{Input}{Input}\SetKwInOut{Output}{Output}
\caption{Sticker-TTS Framework}
\label{alg:sticker_tts_recursive_reasoning}

\Input{Question $Q$, Sticker Extractor $E$, Sticker Modifier $M$, Sticker Utilizer $U$, Max Iterations $N$}
\Output{Final Answer $A_{final}$}

\BlankLine
\textbf{\textcolor[RGB]{0, 130, 0}{// Initial response without sticker}} \\
$T^{(0)}, A^{(0)} \gets U(Q)$  
\Comment{$U$ generate response without sticker}
\\ $TraceList \gets []$, $ AnswerList \gets []$
\\ $TraceList.append(T^{(0)}))$ $AnswerList.append(A^{(0)}))$

\BlankLine
\textbf{\textcolor[RGB]{0, 130, 0}{// Recursive Reasoning Loop}} \\
\For{$k \gets 1$ \KwTo $N$}{
    \textbf{\textcolor[RGB]{0, 130, 0}{// Sticker Extraction}} \\
    $s^{(k)} \gets E(T^{(k-1)}, Q)$ 

    \BlankLine
    \textbf{\textcolor[RGB]{0, 130, 0}{// Sticker Modification}} \\
    $s^{(k)'} \gets M(s^{(k), Q})$ 

    \BlankLine
    \textbf{\textcolor[RGB]{0, 130, 0}{// Trace Generation}} \\
    $T^{(k)}, A^{(k)} \gets U(s^{(k)'}, Q, A^{(k-1)})$ 
    \\ $TraceList.append(T^{(k)})$  $AnswerList.append(A^{(0)}))$
}

\BlankLine
\textbf{\textcolor[RGB]{0, 130, 0}{// Final Answer Derivation}} \\
$A_{final} \gets MajorityVote(AnswerList)$ \Comment{Aggregate answers from all $N$ traces}

\end{algorithm}

To obtain these components, we adopt a two-stage training strategy with distillation-guided self-improvement. At the first training stage, we initialize the framework through knowledge distillation from powerful teacher models. Specifically, we construct training data in the required format by distilling the teacher's reasoning traces, then perform fine-tuning to adapt all three models ($E$, $M$, and $U$) to their respective functional roles. Subsequently, we implement a self-improvement training stage where the framework autonomously generates iterative reasoning traces on open-source mathematical problems. These generated experiences undergo rigorous filtering based on solution validity and reasoning refinement trajectories, forming high-quality self-distilled data. We then conduct additional fine-tuning using this curated dataset to further enhance the models' ability in sticker extraction, error correction, and iterative optimization.


\subsection{Recursive Reasoning Loop}
\label{sec:recursive_reasoning_loop}

Sticker-TTS operates through an iterative mechanism that progressively enhances reasoning quality. As illustrated in Figure~\ref{fig:pipeline}, each iteration $k$ (starting from $k=1$) builds upon the previous reasoning trace $T^{(k-1)}$ and corresponding answer $A^{(k-1)}$. Notably, $T^{(0)}$ denotes the initial response generated by the Sticker Utilizer $U$ without prior sticker integration, and $A^{(0)}$ indicates the answer extracted from $T^{(0)}$. Subsequently, our approach sequentially invokes three phases, \ie sticker extraction, sticker modification, and trace generation, within each iteration, and terminates the generation process utill meeting the stopping criterion.
Below, we formalize the overall recursive process and provide the complete algorithmic flow in Algorithm~\ref{alg:sticker_tts_recursive_reasoning}. 

\begin{promptbox}[Prompt for Sticker Extraction]{lightblue}
Given the solution provided below, Generate an abstract of the key conditions that help solve the problem. The abstract should include both the key conditions and the question.

Abstract Format:

\textbf{Conditions:}

1. [Condition 1]

... (add more conditions as needed)

\textbf{Question:}

[Clearly state what is being asked.]

\textbf{Requirements:}

[Specify requirements that the model must meet.]


   


   

\textbf{Solution to question:}

[Solution]

Please provide your output strictly following ...
\end{promptbox}

\paratitle{Sticker Extraction.} The Sticker Extractor $E$ is designed to effectively capture the primary strategy and reasoning steps while identifying weaknesses in an existing reasoning trace. It takes a reasoning trace $T^{(k-1)}$ and the corresponding question $Q$ as input. Based on this historical trace, $E$ extracts a structured sticker $s^{(k)}$. This sticker acts as a diagnostic summary that captures the strategic essence while pinpointing the most critical limitations within the current reasoning trace. We show the utilized prompt in the following table.

\begin{promptbox}[Prompt for Sticker Modifier]{lightblue}
Given a question and the abstract generated from the solution, carefully check and verify whether the extracted key conditions contain any errors in reasoning or incorrect conditions.

\textbf{ Step 1: Verify and refine the Conditions section.}

- Conditions can come from the reasoning process.



... (Some other requirements are ommited) ...

\textbf{Step 2: Verify the **Question** section.}

- Ensure the question summary is concise ...

- If incorrect, provide a refined version.

\textbf{Step 3: Generate the output. }

- you should output your refined abstract in the following format:
   
  **Conditions:**  
  
  1. [Corrected Condition 1]  
  
  
  ... (more conditions if necessary)  

  **Question:**  
  
  [Refined question summary]  



%

Please provide your output strictly following the step 3 without other unnecessary words.
\end{promptbox}

\paratitle{Sticker Modification.} The Sticker Modifier $M$ examines the sticker $s^{(k)}$ to refine potential errors. According to the reasoning steps and limitations summarized in the sticker, $M$ performs fine-grained error analysis, including computational mistakes and methodological flaws. This process generates a revised sticker $s^{(k)'}$ that incorporates corrective feedback, ensuring subsequent reasoning steps address previously identified weaknesses. We show the utilized prompt in the following table.

\begin{promptbox}[Prompt for Sticker Utilization]{lightblue}
Given a question:

\textbf{[Question]}

Given a sticker that may be correct or incorrect:

\textbf{[Sticker]}

The previous answer that may be correct or incorrect:

\textbf{[Answer]}

Please reason step by step and put final answer in the \\boxed{}.
\end{promptbox}

\paratitle{Sticker Utilization.} The Sticker Utilizer $U$ generate a new reasoning path $T^{(k)}$ by integrating $s^{(k)'}$ with the original question $Q$ and the previous answer $A^{(k-1)}$. The new generated $T^{(k)}$ and $A^{(k)}$ subsequently serve as the input for the next iteration, enabling progressive refinement. We show the utilized prompt in the following table.

\paratitle{Stopping Criterion.} The iterative loop terminates after $N$ iterations, yielding $N$ progressively refined reasoning traces $\{T^{(1)},...,T^{(k)}\}$ and corresponding answers $\{A^{(1)},...,A^{(k)}\}$. To derive the final answer, we aggregate all $N$ answers through the majority vote approach.

\subsection{Self-improvement Progressive Training}  
\label{sec:training}  

Although the design of our framework is clear, developing the framework's components from scratch poses significant challenges, primarily due to the need for a nuanced understanding of complex reasoning patterns. To tackle this issue, we propose a two-stage progressive training strategy. First, we utilize knowledge distillation to align the model with the target inference patterns~(\ie extracting stickers, modifying stickers, and utlizing stickers). Following this, we enhance the model's performance through self-improvement bootstrapping. This approach not only streamlines the training process but also ensures a more robust understanding of the reasoning required for effective performance.


\paratitle{Initialization via Knowledge Distillation.}
The first stage involves training model components to handle complex reasoning via distilled examples from powerful teacher models. We construct task-aligned training data using mathematical problems marked as solvable in the OpenThoughts dataset~\cite{Open-Thoughts} and employ powerful \texttt{DeepSeek-R1}~\cite{guo2025deepseek} to generate high-quality reasoning traces. For training Sticker Extractor $E$, we use \texttt{o3-mini}~\cite{openai2025o3} to extract structured stickers from the long-form reasoning traces, which exhibit greater faithfulness compared to other reasoning models~\cite{hhem-2.1-open}. Subsequently, to prepare training data for models Sticker Modifier $M$ and Sticker Utilizer $U$, we simulate error-correction scenarios. Specifically, we start from the flawed reasoning traces and their corresponding stickers derived from the training data prepared for Sticker Extractor and leverage \texttt{DeepSeek-R1} as Sticker Modifier and Utilizer to examine stickers and generate refined reasoning paths. Finally, We only retain the generated data from the three models to form paired training data on the condition that the final reasoning trajectory is completely correct. Through fine-tuning on these distilled datasets, each component has preliminarily acquired its specialized capability in extraction, correction, and optimization.



\paratitle{Self-improvement Bootstrapping.}
To further enhance the model's capabilities, we enable the model to generate data autonomously and employ rigorous curation of the self-distilled training data.
Leveraging the initialized framework, we iteratively generate reasoning traces on OpenThoughts while enforcing dual filtering criteria to ensure the quality of the training data. The first criterion is \textit{solution validity}. We preserve trajectories where the final optimized answer is correct while maintaining a 1:2 ratio between ``error-to-correct'' transitions (where the initial reasoning path contains errors but the final optimized answer is corrected) and ``correct-to-correct'' transitions (where the training path is already valid while undergoing further refinement). This ratio aligns with the statistical distribution of naturally generated reasoning paths, where correct initial attempts occur more frequently. The second criterion is \textit{correction significance}. For selected cases where iterative refinement succeeds after previous reasoning fails, we limit the preceding two iterations to yield incorrect answers. This ensures the difficulty of the retained cases, which involve non-trivial corrections requiring sustained reasoning effort.
Subsequent fine-tuning on this curated dataset enables synergistic enhancement of the framework: Sticker Extractor $E$ improves its capacity to identify critical reasoning patterns from iterative histories, Sticker Modifier $M$ develops robust error diagnosis through exposure to multi-failure recovery scenarios, and Sticker Utilizer $U$ strengthens its reasoning path generation capability by integrating optimized reasoning strategies. 
To prevent overfitting, we limit each mathematical problem to provide at most one qualified training instance for each framework component during their respective training phases.

\section{Experiments}
\label{sec:experiments}

\begin{table*}[h]
\centering
\resizebox{\textwidth}{!}{ 
\begin{tabular}{lccccccccc}
\toprule
\multirow{2}{*}{\textbf{Method}} & \multicolumn{3}{c}{\textbf{AIME 2024}} & \multicolumn{3}{c}{\textbf{AIME 2025}} & \multicolumn{3}{c}{\textbf{OlymMATH-EN-EASY}} \\
\cmidrule(lr){2-4} \cmidrule(lr){5-7} \cmidrule(lr){8-10}
 & \textbf{Pass@1} & \textbf{Cons@20} & \textbf{Cons@64} & \textbf{Pass@1} & \textbf{Cons@20} & \textbf{Cons@64} & \textbf{Pass@1} & \textbf{Cons@20} & \textbf{Cons@64} \\
\midrule
\multicolumn{10}{l}{\textit{7B Models}} \\
\cmidrule(lr){1-10}
DeepSeek-R1-Distill & 55.52 & 73.33 & 76.67 & 38.54 & 53.33 & 56.67 & 41.88 & 67.00 & 71.00 \\
\cmidrule(lr){1-10}
Light-R1 & 57.55 & 76.67 & 80.00 & 42.86 & 53.33 & 60.00 & 46.48 & 65.00 & 74.00 \\
Skywork-OR1 & \underline{66.30} & 76.67 & 83.33 & \textbf{52.50} & \textbf{63.33} & 63.33 & 57.38 & \underline{79.00} & 78.00 \\
\cmidrule(lr){1-10}
Think-Twice & 56.67 & 73.33 & 76.67 & 43.33 & 56.67 & 56.67 & 53.00 & 55.00 & 58.00 \\
LeaP-T & 64.38 & \underline{80.00} & 80.00 & 41.25 & 56.67 & 60.00 & 35.95 & 62.00 & 68.00 \\
\cmidrule(lr){1-10}
Ours (Stage 1, $N$=10) & 60.00 & \underline{80.00} & / & 40.00 & \underline{60.00} & / & \underline{61.00} & 76.00 & / \\
Ours (Stage 2, $N$=10) & \textbf{66.67} & \textbf{83.33} & / & \underline{43.33} & \textbf{63.33} & / & \textbf{63.00} & \textbf{80.00} & / \\
\midrule
\multicolumn{10}{l}{\textit{32B Models}} \\
\cmidrule(lr){1-10}
DeepSeek-R1-Distill & 72.60 & 83.33 & 86.67 & 54.37 & 70.00 & 73.33 & 65.34 & 86.00 & 87.00 \\
\cmidrule(lr){1-10}
Light-R1 & 76.77 & 86.67 & 86.67 & 64.79 & 73.33 & 76.67 & 75.53 & 89.00 & 92.00 \\
Skywork-OR1 & \underline{80.83} & 86.67 & 86.67 & 72.08 & \underline{80.00} & 80.00 & \underline{85.77} & \underline{93.00} & 96.00 \\
AM-Thinking-v1 & \textbf{81.15} & \underline{90.00} & 90.00 & \textbf{76.25} & \textbf{83.33} & 83.33 & \textbf{86.25} & \textbf{95.00} & 96.00 \\
\cmidrule(lr){1-10}
Ours (Stage 1, $N$=10) & 70.00 & \underline{90.00} & / & 70.00 & \underline{80.00} & / & 79.00 & 88.00 & / \\
Ours (Stage 2, $N$=10) & 76.67 & \textbf{93.33} & / & \underline{73.33} & \underline{80.00} & / & 78.00 & 90.00 & / \\
\bottomrule
\end{tabular}
}
\caption{Evaluation results on three mathematical reasoning benchmarks. Note that while our method reports answers via Cons@N, its associated reasoning cost is comparable to Cons@2N. To ensure fair comparison, performance comparisons are conducted with aligned reasoning consumption. We additionally provide baseline reference performance at larger $N$ values for context. The best and second-best results are highlighted in \textbf{bold} and \underline{underlined}, respectively.}
\label{tab:results}
\end{table*}

\subsection{Experimental Setup}
\label{sec:experimental_setup}

\paratitle{Dataset and Benchmarks.}  We evaluate our method on three mathematical reasoning benchmarks: AIME 2024~\cite{aime2024}, AIME 2025~\cite{aime2025}, and OlymMATH-EN-EASY~\cite{olymmath-paper}. AIME offers $30$ challenging mathematical problems per year targeting academically advanced high school students. OlymMATH-EN-EASY comprises $100$ Olympiad-level problems, designed to rigorously evaluate complex reasoning capabilities with verifiable numerical solutions. For model training, we use the math subset of OpenThoughts~\cite{Open-Thoughts}, an open synthetic reasoning dataset containing $114k$ high-quality examples.

\paratitle{Evaluation Metrics.}  We employ two primary metrics: Pass@1 and Cons@N.
For baseline models, Pass@1 is estimated by generating $64$ responses per query using nucleus sampling with a top-$p$ value of $0.95$ and a temperature of $0.6$. In our method, Pass@1 is computed directly using the answer from the final iteration.
Cons@N evaluates the majority vote agreement, where baseline implementations generate $N$ independent samples, while our method naturally accumulates $N$ responses through iterations and performs voting across these evolution trajectories.
To ensure fair comparison, we configure generation parameters consistently across models. For the \texttt{DeepSeek-R1-Distill-Qwen}\footnote{\url{https://huggingface.co/deepseek-ai/DeepSeek-R1-Distill-Qwen-32B}} series, the maximum generation length is set to $32,000$ tokens. For the \texttt{Qwen2.5} series~\cite{qwen2.5-report}, the maximum generation length is configured to $5,000$ tokens.

\paratitle{Baselines.} To ensure comprehensive evaluation, we consider LLMs trained via three approaches as baselines, including \textit{distillation}, \textit{multi-staged post-training featuring RL}, and \textit{test-time scaling framework}. For the distillation approach, we adopt the \texttt{DeepSeek-R1-Distill} series as evaluation baselines. For the multi-staged post-training method with RL, we employ the \texttt{Light-R1} series~\cite{lightr1proj}, \texttt{Skywork-OR1} series~\cite{skywork-or1-2025}, and \texttt{AM-Thinking-v1}~\cite{ji2025thinking} as baseline LLMs. For the test-time scaling framework approach, we utilize \texttt{LeaP-T-7B}~\cite{luo2025learning} and Think-Twice~\cite{tian2025think} as baselines.

\paratitle{Implementation Details.} For data preparation, we employ \texttt{DeepSeek-R1-Distill-Qwen-7B} to sample $10$ reasoning trajectories per mathematical problem in the OpenThoughts dataset. The correctness rates of these trajectories are used to estimate problem difficulty levels. During the knowledge distillation stage, we select problems with difficulty scores between $0.2$ and $0.5$. Responses from \texttt{DeepSeek-R1} and \texttt{o3-mini} are obtained via API calls, with sampling parameters the same as the evaluation setup. For the self-improvement bootstrapping stage, we curate more challenging data with difficulty scores ranging from $0$ to $0.4$.
The Sticker Extractor is trained using the \texttt{Qwen2.5} series models, while both the Sticker Modifier and Sticker Utilizer utilize the \texttt{DeepSeek-R1-Distill-Qwen} series. Experiments are conducted across two model scales: 7B and 32B. As for the SFT configuration, the maximum context length is $20,000$ tokens. The Sticker Extractor is trained with a batch size of $96$ and a learning rate of $1 \times 10^{-5}$. The Sticker Modifier and Sticker Utilizer are trained with a batch size of $128$ and a learning rate of $2 \times 10^{-5}$. The detailed information of SFT configuration is in Appendix~\ref{appendix:sft}.

\subsection{Main Results}
\label{sec:main_results}

Table~\ref{tab:results} presents the performance of our method and other baselines on three representative mathematical reasoning datasets. We can make the following observations:

$\bullet$ \emph{Superior Performance.} Our proposed method demonstrates better or comparable performance compared to other baselines. After the first training stage of distillation, our method already surpasses models trained through distillation and test-time scaling framework. Following the second training stage, our method outperforms most baselines, and even exceeds some models developed via multi-staged post-training featuring RL, such as \texttt{Light-R1}. Its performance is comparable to the current state-of-the-art open-source reasoning model \texttt{AM-Thinking-v1} with the metric of  Cons@20. This two-stage progression indicates that the initial knowledge distillation successfully adapts the framework's components to their functional roles, while the subsequent self-improvement bootstrapping enables synergistic capability enhancement of the framework. The sustained performance gains confirm our framework's powerful capacity for reasoning path optimization and generation.

$\bullet$ \emph{Scalability Across Model Sizes.} Our method demonstrates effectiveness across different model scales, achieving considerable improvements with both 7B and 32B parameter variants. This scalability demonstrates that our framework adapts well to varying model capability levels. Our framework enables effective division of labor regardless of base model size, with each component specializing in its respective task while maintaining coherent collaboration.


$\bullet$ \emph{Enhanced Reasoning Efficiency}. Our method achieves substantial performance with a favorable reasoning cost. We approximate inference cost by the number of long-CoT solutions generated, as token generation by models (\eg \texttt{DeepSeek-R1-Distill-Qwen} series) is the primary bottleneck in deployment. This efficiency stems from our framework's architecture: ``stickers'' generated by the Sticker Extractor eliminate lengthy reasoning traces and thus have minimal overhead, while the Sticker Modifier and Optimizer are the main long-CoT components. Consequently, the total reasoning cost for $N$ iterations is comparable to generating $2N$ long-CoT solutions. This allows our method with only $N$=10 iterations (at a cost comparable to Cons@20) to outperform the Cons@64 performance of most baselines. For a fair comparison, baselines like Think-Twice that leverage prior answers are also assessed over $2N$ rounds, aligning their costs. Additionally, as illustrated in Table~\ref{tab:cons_n_performance}, our model achieves effective test-time scaling as $N$ increases because each iteration effectively distills insights from prior attempts instead of conducting history-unaware parallel sampling.

\begin{table}[t]
\centering
\small
\resizebox{0.4\textwidth}{!}{
\begin{tabular}{cc}
\toprule
\textbf{Iteration Number ($N$)} & \textbf{Cons@N (\%)} \\
\midrule
2  & 56.67 \\
4  & 73.33 \\
8  & 80.00 \\
16 & 83.33 \\
\bottomrule
\end{tabular}
}
\caption{Cons@N on OlymMATH-EN-EASY across varying iteration number $N$ of 7B model.}
\label{tab:cons_n_performance}
\end{table}




\begin{table}[t]
\centering
\small
\resizebox{0.5\textwidth}{!}{ 
\begin{tabular}{lcc}
\toprule
\textbf{Method} & \makecell{\textbf{AIME 2024} \\ \textbf{(Cons@20)}} & \makecell{\textbf{AIME 2025} \\ \textbf{(Cons@20)}} \\
\midrule
\multicolumn{3}{l}{\textit{7B Models}} \\
\cmidrule(lr){1-3}
Ours (Stage 2, $N$=10) & 83.33 & 63.33 \\
Early Exit & 70.00 & 56.67 \\
Parallel Sampling ($P$=2, $Q$=5) & 80.00 & 60.00 \\
Parallel Sampling ($P$=5, $Q$=2) & 80.00 & 56.67 \\
\midrule
\multicolumn{3}{l}{\textit{32B Models}} \\
\cmidrule(lr){1-3}
Ours (Stage 2, $N$=10) & 93.33 & 80.00 \\
Early Exit & 86.67 & 76.67 \\
Parallel Sampling ($P$=2, $Q$=5) & 93.33 & 73.33 \\
Parallel Sampling ($P$=5, $Q$=2) & 86.67 & 76.67 \\
\bottomrule
\end{tabular}
}
\caption{Performance comparison under different iteration strategies.}
\label{tab:reason-depth}
\end{table}

\begin{table}[t]
\centering
\small
\resizebox{0.46\textwidth}{!}{ 
\begin{tabular}{lcc}
\toprule
\textbf{Method} & \makecell{\textbf{AIME 2024} \\ \textbf{(Cons@20)}} & \makecell{\textbf{AIME 2025} \\ \textbf{(Cons@20)}} \\
\midrule
\multicolumn{3}{l}{\textit{7B Models}} \\
\cmidrule(lr){1-3}
Ours (Stage 2, $N$=10) & 83.33 & 63.33 \\
Extractor Ablation & 73.33 & 53.33 \\
Modifier Ablation & 70.00 & 53.33 \\
Full Ablation & 70.00 & 50.00 \\
\bottomrule
\end{tabular}
}
\caption{Ablation study in Sticker-TTS.}
\label{tab:ablation}
\end{table}

\begin{table}[t]
\centering
\small
\resizebox{0.5\textwidth}{!}{ 
\begin{tabular}{lcccc}
\toprule
\multirow{2}{*}{\textbf{Method}} & \multicolumn{2}{c}{\textbf{AIME 2024}} & \multicolumn{2}{c}{\textbf{AIME 2025}} \\
\cmidrule(lr){2-3} \cmidrule(lr){4-5}
 & \textbf{Pass@1} & \textbf{Cons@20} & \textbf{Pass@1} & \textbf{Cons@20} \\
\midrule
\multicolumn{5}{l}{\textit{32B Models}} \\
\cmidrule(lr){1-5}
DeepSeek-R1-Distill & 72.60 & 83.33 & 54.37 & 70.00 \\
Light-R1 & 76.77 & 86.67 & 64.79 & 73.33 \\
Sticker Utilizer & 75.68 & 86.67 & 58.54 & 73.33 \\
\bottomrule
\end{tabular}
}
\caption{Evaluation results of the 32B Sticker Utilizer.}
\label{tab:sticker-optimizer}
\end{table}

\subsection{Further Analysis}
\label{sec:further_analysis}

\paratitle{Reasoning Depth}. Since our method continually refines its outputs by leveraging the history of prior responses, we can vary the number of iterations $N$ to control the reasoning depth.
We examine two strategies: \textit{early exit} and \textit{parallel sampling}. For early exit, an additional stopping criterion is introduced where the iteration terminates if the current response's answer matches that of the previous iteration. For parallel sampling, we partition the sampling process into $P$ parallel chains, each executing $Q$ iterations per query, ensuring $PQ=N$. The results of these experiments are presented in Table~\ref{tab:reason-depth}. Overall, we can have two major observations. Firstly, increasing test time enables our method to better learn from experience. While the early exit strategy reduces the average number of iterations, it appears detrimental to the refinement of stickers through deeper iterations, thereby limiting the depth of perception and learning from historical responses. Secondly, with the same reasoning costs, deeper iterations yield consistent performance gains over other methods, indicating that our method effectively leverages historical responses for sustained optimization. This suggests that the interplay among the three Sticker components progressively strengthens the consensus and accuracy of the reasoning outcome.

\paratitle{Ablation Study.} To assess the effectiveness of components in our framework, we conduct ablation experiments focusing on the Sticker Extractor and Sticker Modifier. Three configurations are tested: (1) \textit{Extractor Ablation}: Directly feeding raw reasoning traces to the Sticker Modifier without sticker extraction; (2) \textit{Modifier Ablation}: Using unmodified stickers from the Extractor to generate new traces; (3) \textit{Full Ablation}: Generating new traces directly from the original reasoning path without sticker involvement. As shown in Table~\ref{tab:ablation}, performance declines under individual component ablation, while full ablation causes the most significant degradation. This demonstrates that both components serve critical roles: the Sticker Extractor's strategy abstraction prevents the Sticker Modifier from being overwhelmed by details in reasoning traces, while the Sticker Modifier's error correction ensures sticker quality for subsequent optimization. The compounded performance loss under full ablation suggests that intermediate sticker representations are likely essential for navigating the internal complexity of reasoning traces. Without structured stickers, the framework struggles to maintain strategic focus during iterative refinement, potentially propagating errors or becoming trapped in flawed reasoning patterns.

\paratitle{Sticker Utilizer Analysis.} We conduct standalone evaluations of the 32B Sticker Utilizer after the two-stage training, without the collaboration of the other two models. As shown in Table~\ref{tab:sticker-optimizer}, the Sticker Utilizer achieves superior performance compared to \texttt{DeepSeek-R1-Distill-32B} while matching \texttt{Light-R1} in Cons@20 metrics. This demonstrates that training models to optimize reasoning paths enhances intrinsic reasoning capabilities. Notably, while the Sticker Utilizer's Pass@1 score is lower than \texttt{Light-R1}, likely due to differences in training objectives, its Cons@20 equivalence shows that the majority vote strategy effectively overcomes the instability of single run by aggregating diverse valid trajectories. This suggests that the Sticker Utilizer possesses strong reasoning potential, and its generation stability could be enhanced with further calibration.

\section{Related Work}
\label{sec:related_work}

\paratitle{Test-Time Scaling Techniques.}
Recent advances have proposed a range of decoding strategies to enhance reasoning accuracy during inference. A prominent line of work involves performing multiple sampling passes and selecting the final answer via majority voting, as exemplified by the self-consistency method~\cite{wang2022self}. Building on this, confidence-weighted self-consistency~\cite{taubenfeld2025confidence} reduces the number of required samples by incorporating answer uncertainty. Beyond independent sampling, recent approaches leverage multiple rounds of generation informed by previous attempts, such as feeding the full prior answer back into the model~\cite{tian2025think} or adopting parallel thinking mechanisms~\cite{luo2025learning}. However, these long-form reasoning processes impose a significant burden on the model's long-context capabilities~\cite{li2023loogle}, while overly brief answers limit the potential to leverage historical attempts effectively. To mitigate this burden, AOT~\cite{aot-fengwei} structures reasoning as a Markov process, iteratively decomposing and contracting problems into independent atomic units to eliminate reliance on historical information. Moreover, existing methods primarily focus on prompt design and offer limited support for iterative improvement through training. In contrast, our proposed framework introduces \textit{stickers}, which are succinct, distilled cues extracted from extended reasoning traces, to guide the utilization of historical solutions. Furthermore, complemented by a two-stage training strategy that combines imitation learning and self-improvement, our framework enables continual enhancement of test-time reasoning performance.

\paratitle{Reinforcement Learning for Reasoning.} 
With the help of RL, LRMs have achieved significant progress. Especially, OpenAI’s o1 series~\footnote{https://openai.com/o1}, DeepSeek-R1~\cite{guo2025deepseek}, and Kimi K1.5~
\cite{team2025kimi} have achieved surprising math and code performance by training with outcome-based reward on large scale. Complementary to this, methods like VC-PPO~\cite{yuan2025vcppo}, and Light-R1~\cite{wen2025light} investigate alternative reward formulations, curriculum learning, and multi-stage training to enhance reasoning capabilities. The proliferation of open-source frameworks—including SimpleRL~\cite{zeng2025simplerl} and STILL series work~\cite{chen2025empirical}-has played a vital role in replicating and scaling RL pipelines, promoting reproducibility and accelerating broader adoption. These advances collectively provide a robust foundation for efficient and reliable RL training in large models. Our approach is decoupled from the underlying model, making it pluggable with the aforementioned models to enhance their test-time scalability and performance. Additionally, our training strategy can be applied to further improve the overall performance.

\section{Conclusion}
In this paper, we explore how to enhance the test-time scaling performance of LRMs. We propose a novel sticker-based test-time scaling framework which consists of three modules: a \textit{Sticker Extractor} to distill concise and relevant insights (``stickers'') from previous reasoning traces; a \textit{Sticker Modifier} to adapt these stickers to the current context; and a \textit{Sticker Utilizer} to integrate them to guide the model towards more effective solution strategies. During inference, these components operate iteratively, allowing the model to synthesize prior knowledge with new reasoning paths. Extensive experiments validate its effectiveness, demonstrating its superiority over strong baselines.

\section*{Acknowledgment}
This work was partially supported by National Natural Science Foundation of China under Grant No. 92470205 and 62222215, Beijing Natural Science Foundation under Grant No. L233008 and Beijing Municipal Science and Technology Project under Grant No. Z231100010323009. Xin Zhao is the corresponding author.

\section*{Limitations}
In this paper, we present a sticker-based test-time scaling framework to enhance reasoning capacities of LRMs during inference. Beyond the \texttt{DeepSeek-R1-Distill-Qwen} model, we believe our framework can be employed in broader LRMs, which have not been explored owing to computational costs. Additionally, our method mainly focuses on utilizing supervised fine-tuning~(\ie RFT) to train each module in the framework. A key challenge in such a multi-stage pipeline is the potential for error propagation. For instance, failures in the initial sticker extraction or the encoding of spurious correlations could potentially steer the iterative process into unproductive loops, particularly on ill-structured tasks. While our framework mitigates this through built-in safeguards—a dedicated error-correction module (Sticker Modifier), robust aggregation via majority vote, and a training process focused on failure recovery—a more extensive failure analysis remains valuable. In the future, we can further employ RL to train the whole framework, which is a multi-agent system in essence. Limited by computational costs, we conduct experiments on models up to 32B in size, and the method's generalization to broader domains requires further exploration. Consequently, our future work may explore validating our proposed method on even larger models and investigate test-time scaling techniques to enhance its domain generalization capabilities.

\bibliography{acl_latex}

\appendix

\clearpage
\appendix
\section{SFT Configuration}
\label{appendix:sft}

We utilize the huggingface Transformers~\cite{wolf-2019-huggingface} to implement our experiments, using Flash Attention~\cite{flash_attn} and DeepSpeed ZeRO Stage 3 to optimize the training efficiency. We employ AdamW optimizer~\cite{adamw} with $\beta_1=0.9$ and $\beta_2=0.95$, and use the cosine learning rate scheduler. We use BFloat16 mixed precision, with a warmup ratio of $0.1$ and a weight decay of $0.1$ to ensure training stability. To enhance computational efficiency, we apply gradient checkpointing strategy~\cite{chen-2016-training}.

\section{Prompt Engineering}
This section details the specific prompts used for the three core components of the Sticker-TTS framework: the Sticker Extractor, Sticker Modifier, and Sticker Utilizer.

\begin{promptbox}[Prompt for Sticker Extraction]{lightblue}
Given the solution provided below, Generate an abstract of the key conditions that help solve the problem. The abstract should include both the key conditions and the question.

Abstract Format:

**Conditions:**

1. [Condition 1]

2. [Condition 2]

... (add more conditions as needed)

**Question:**

[Clearly state what is being asked.]

Requirements:

1. **Conditions**

   - Only retain the key steps that directly impact solving the problem, ignoring lengthy derivations and irrelevant calculations.
   
   - Each step must have a clear mathematical significance, meaning it makes a substantial contribution to the final conclusion.
   
   - The conditions can come from the reasoning process.
   
   - Write each condition on a separate line, numbered sequentially.
   
   - Remove repetitive calculations and obvious equation transformations.
   
   - **List as many conditions as possible**
   
   - **Do not record direct substitutions of common formulas unless they play a key role in the derivation.**
   
   - **Each condition must be atomic and indivisible** (i.e., it cannot be divided into two sub-conditions).
   
   - **Each condition must refer to something clearly and without ambiguity.**!!!

2. **Question:**

   - Summarize what is being asked in one clear sentence.
   
   - Remove all known conditions.

solution to question:

\{solution\}

Please provide your output strictly following the abstract format without other unnecessary words.
\end{promptbox}

\begin{promptbox}[Prompt for Sticker Modifier]{lightblue}
Given a question and the abstract generated from the solution, carefully check and verify whether the extracted key conditions contain any errors in reasoning or incorrect conditions.

\#\#\# Step 1: Verify and refine the **Conditions** section.

- **Conditions can come from the reasoning process.**

- Check if any condition includes unnecessary reasoning or incorrect logic. If it exists, it must be refined.

- Ensure all conditions are atomic and indivisible.

- Ensure all conditions must refer to something clearly and without ambiguity.

- If a condition is derived through reasoning, please strictly verify whether it is correct and contributes to solving the problem. If there is a problem, refine it.

- If any key condition is missing or incorrectly formulated, supplement or refine it.

\#\#\# Step 2: Verify the **Question** section.

- Ensure the question summary is concise and does not include any known conditions.

- If incorrect, provide a refined version.

\#\#\# Step 3: Generate the output. 

- you should output your refined abstract in the following format:
   
  **Conditions:** 1. [Corrected Condition 1]  
  
  2. [Corrected Condition 2]  
  
  ... (more conditions if necessary)  

  **Question:** [Refined question summary]  

Here is the given question:

\{question\}

Here is the given abstract:

\{sticker\}

Please provide your output strictly following the step 3 without other unnecessary words.
\end{promptbox}

\begin{promptbox}[Prompt for Sticker Utilization]{lightblue}
Given a question:

\{question\}

Given a sticker that may be correct or incorrect:

\{sticker\}

The previous answer that may be correct or incorrect:

\{answer\}

Please reason step by step and put the final answer in the $\arraybackslash$ boxed\{\}.
\end{promptbox}








\end{document}